\documentclass[runningheads]{llncs}
\usepackage[T1]{fontenc}
\usepackage{graphicx}
\usepackage{amsmath}
\usepackage{amssymb}
\usepackage{booktabs}
\usepackage[table,xcdraw]{xcolor}
\usepackage{threeparttable}
\usepackage{amssymb}
\usepackage{tabularx}
\usepackage{xcolor}
\usepackage{colortbl}
\definecolor{mygray}{gray}{.9}
\usepackage{booktabs}
\usepackage{wrapfig} 
\usepackage{caption}
\usepackage[colorlinks=true,citecolor=blue,urlcolor=blue]{hyperref}
\makeatletter\renewcommand\paragraph{\@startsection{paragraph}{4}{\z@}
  {.5em \@plus1ex \@minus.2ex}{-.5em}{\normalfont\normalsize\bfseries}}\makeatother
 
\begin{document}
\title{SwinMM: Masked Multi-view with Swin Transformers for 3D Medical Image Segmentation}
\titlerunning{SwinMM}
%

\author{Yiqing Wang\inst{1}\thanks{Equal contribution} \and
Zihan Li\inst{2}$^{\star}$ \and
Jieru Mei\inst{3}$^{\star}$ \and
Zihao Wei\inst{1,7}$^{\star}$ \and
Li Liu\inst{4} \and
Chen Wang\inst{5} \and
Shengtian Sang\inst{6} \and
Alan Yuille\inst{3} \and
Cihang Xie\inst{4} \and
Yuyin Zhou\inst{4} 
}

\authorrunning{Y. Wang et al.}

\institute{Shanghai Jiao Tong University \and
University of Washington \and
The Johns Hopkins University \and
University of California, Santa Cruz \and
Tsinghua University \and
Stanford University \and
University of Michigan, Ann Arbor
}
\maketitle              
\begin{abstract}

Recent advancements in large-scale Vision Transformers have made significant strides in improving pre-trained models for medical image segmentation. However, these methods face a notable challenge in acquiring a substantial amount of pre-training data, particularly within the medical field. To address this limitation, we present \textbf{M}asked \textbf{M}ulti-view with \textbf{Swin} Transformers (\textbf{SwinMM}), a novel multi-view pipeline for enabling accurate and data-efficient self-supervised medical image analysis.
Our strategy harnesses the potential of multi-view information by incorporating two principal components.  In the pre-training phase, we deploy a masked multi-view encoder devised to concurrently train masked multi-view observations through a range of diverse proxy tasks. These tasks span image reconstruction, rotation, contrastive learning, and a novel task that employs a mutual learning paradigm. This new task capitalizes on the consistency between predictions from various perspectives, enabling the extraction of hidden multi-view information from 3D medical data. In the fine-tuning stage, a cross-view decoder is developed to aggregate the multi-view information through a cross-attention block.
Compared with the previous state-of-the-art self-supervised learning method Swin UNETR, SwinMM demonstrates a notable advantage on several medical image segmentation tasks.
It allows for a smooth integration of multi-view information, significantly boosting both the accuracy and data-efficiency of the model. Code and models are available at \href{https://github.com/UCSC-VLAA/SwinMM/}{https://github.com/UCSC-VLAA/SwinMM/}.

\end{abstract}

\section{Introduction}
Medical image segmentation is a critical task in computer-assisted diagnosis, treatment planning, and intervention. While large-scale transformers have demonstrated impressive performance in various computer vision tasks~\cite{dosovitskiy2020image,kim2022vit,Hatamizadeh2022UNETRTF}, such as natural image recognition, detection, and segmentation~\cite{chen2021transunet,li2022lvit}, they face significant challenges when applied to medical image analysis. The primary challenge is the scarcity of labeled medical images due to the difficulty in collecting and labeling them, which requires specialized medical knowledge and is time-consuming~\cite{tajbakhsh2020embracing,label_difficulty,hong2023distance}. 
The second challenge is the ability to identify sparse and obscure patterns in medical images, including blurred and dim images with small segmentation targets. 
Hence, it is imperative to develop a precise and data-efficient pipeline for medical image analysis networks to enhance their accuracy and reliability in computer-assisted medical diagnoses.

Self-supervised learning, a technique for constructing feature embedding spaces by designing pretext tasks, has emerged as a promising solution for addressing the issue of label deficiency. One representative methodology for self-supervised learning is the masked autoencoder (MAE)~\cite{He2022MaskedAA}. MAEs learn to reconstruct input data after randomly masking certain input features. This approach has been successfully deployed in various applications, including image denoising, text completion, anomaly detection, and feature learning. In the field of medical image analysis, MAE pre-training has also been found to be effective~\cite{Zhou2022SelfPW}. Nevertheless, these studies have a limitation in that they require a large set of unlabeled data and do not prioritize improving output reliability, which may undermine their practicality in the real world.

In this paper, we propose \textbf{M}asked \textbf{M}ulti-view with \textbf{Swin} (SwinMM), the first comprehensive multi-view pipeline for self-supervised medical image segmentation.
We draw inspiration from previous studies~\cite{zhao2022mmgl,zhou2019semi,xia2020uncertainty,zhai2022mvcnet} and aim to enhance output reliability and data utilization by incorporating multi-view learning into the self-supervised learning pipeline.
During the pre-training stage, the proposed approach randomly masks 3D medical images and creates various observations from different views. A masked multi-view encoder processes these observations simultaneously to accomplish four proxy tasks: image reconstruction, rotation, contrastive learning, and a novel proxy task that utilizes a mutual learning paradigm to maximize consistency between predictions from different views. This approach effectively leverages hidden multi-view information from 3D medical data and allows the encoder to learn enriched high-level representations of the original images, which benefits the downstream segmentation task.
In the fine-tuning stage, different views from the same image are encoded into a series of representations, which will interact with each other in a specially designed cross-view attention block. 
A multi-view consistency loss is imposed to generate aligned output predictions from various perspectives, which enhances the reliability and precision of the final output.
The complementary nature of the different views used in SwinMM results in higher precision, requiring less training data and annotations, which holds significant potential for advancing the state-of-the-art in this field. 
In summary, the contributions of our study are as follows:
\begin{itemize}
\item We present SwinMM, a unique and data-efficient pipeline for 3D medical image analysis, providing the first comprehensive multi-view, self-supervised approach in this field.
\item Our design includes a masked multi-view encoder and a novel mutual learning-based proxy task, facilitating effective self-supervised pretraining.
\item We incorporate a cross-view decoder for optimizing the utilization of multi-view information via a cross-attention block.
\item SwinMM delivers superior performance with an average Dice score of 86.18\% on the WORD dataset, outperforming other leading segmentation methods in both data efficiency and segmentation performance.
\end{itemize}

\section{Method}
Figure \ref{framework} provides an overview of SwinMM, comprising a masked multi-view encoder and a cross-view decoder. SwinMM creates multiple views by randomly masking an input image, subsequently feeding these masked views into the encoder for self-supervised pre-training. In the fine-tuning stage, we architect a cross-view attention module within the decoder. This design facilitates the effective utilization of multi-view information, enabling the generation of more precise segmentation predictions.

\subsection{Pre-training}
\label{pre-train}

\begin{figure}[!htb]
    \centering
    \begin{minipage}[t]{0.48\linewidth}
    \setlength{\leftskip}{10pt}
\setlength{\rightskip}{10pt}
        \centering\includegraphics[width=\linewidth]{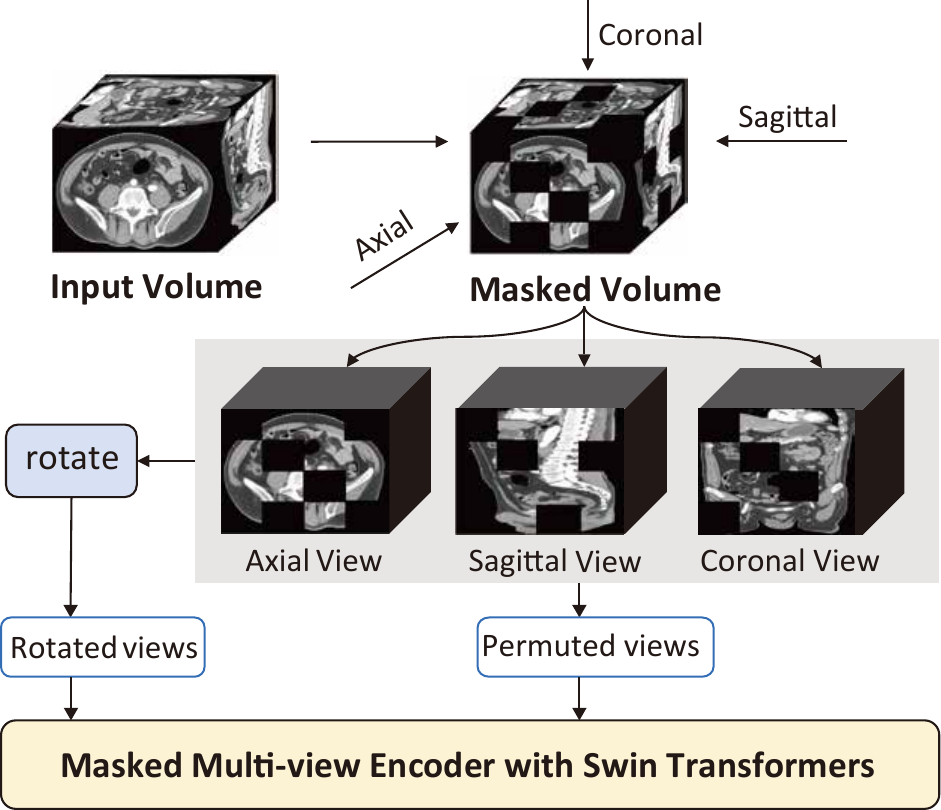}
	\caption{SwinMM's pre-training stage.}
        \label{pretrain}
    \end{minipage}
    \hfill
    \begin{minipage}[t]{0.46\linewidth}
	\centering      \includegraphics[width=\linewidth]{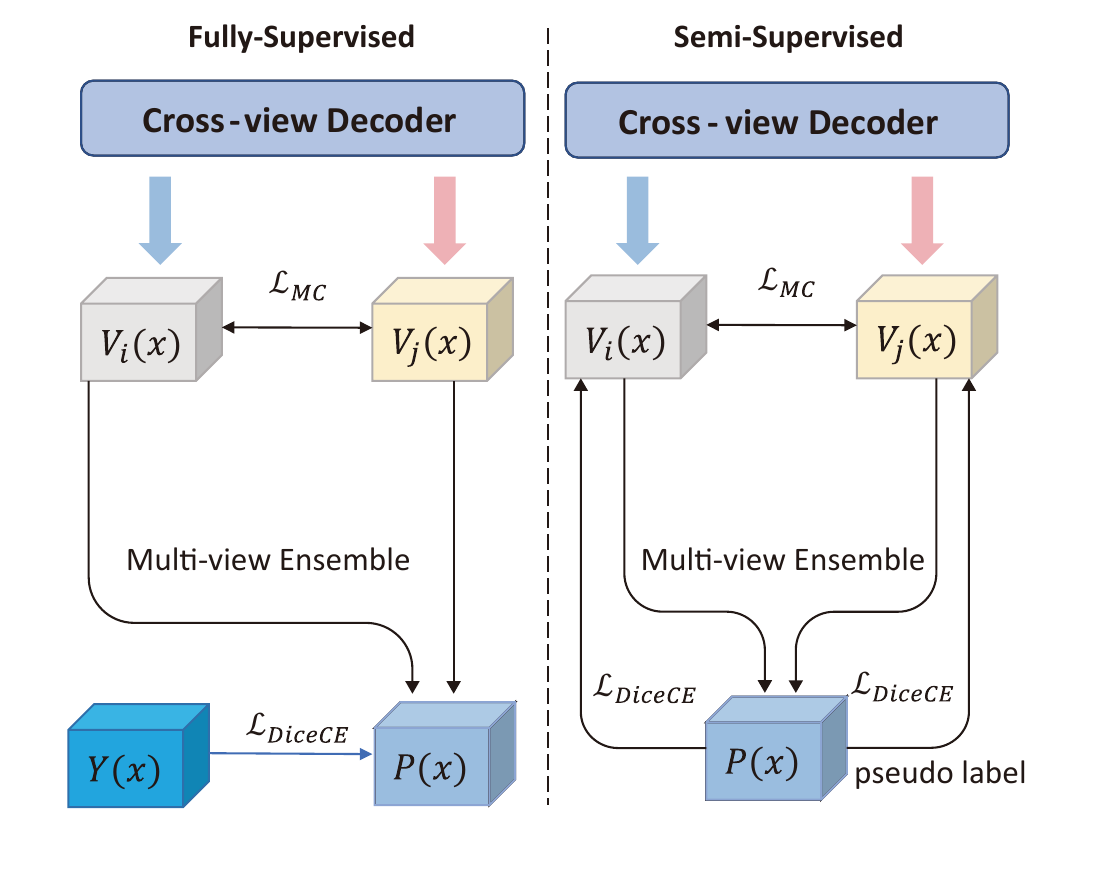}
	\caption{The Fully-supervised / Semi-supervised pipeline with SwinMM.} 
        \label{Semi}
    \end{minipage}
  \setlength{\belowcaptionskip}{-1cm}
\end{figure}

\paragraph{\textbf{Masked multi-view encoder.}}
Following~\cite{He2022MaskedAA}, we divided the 3D images into sub-volumes of the same size and randomly masked a portion of them, as demonstrated in Figure \ref{pretrain}. 
These masked 3D patches, from different perspectives, were then utilized for self-supervised pretraining by the masked multi-view encoder. As shown in Figure \ref{framework}, the encoder is comprised of a patch partition layer, a patch embedding layer, and four Swin Transformer layers~\cite{liu2021swin}. Notably, unlike typical transformer encoders, our masked multi-view encoder can process multiple inputs from diverse images with varying views, making it more robust for a broad range of applications.

\paragraph{\textbf{Pre-training strategy.}}
\label{Pre-training Strategy}
To incorporate multiple perspectives of a 3D volume, we generated views from different observation angles, including axial, coronal, and sagittal. Furthermore, we applied rotation operations aligned with each perspective, consisting of angles of $0^\circ$, $90^\circ$, $180^\circ$, and $270^\circ$ along the corresponding direction.
To facilitate self-supervised pre-training, we devised four proxy tasks. The reconstruction and rotation tasks measure the model's performance on each input individually, while the contrastive and mutual learning tasks enable the model to integrate information across multiple views. 
\begin{itemize}
\item\textbf{The reconstruction task} compares the difference between unmasked input $\mathcal{X}$ and the reconstructed image $y^{rec}$. Following ~\cite{He2022MaskedAA}, we adopt Mean-Square-Error(MSE) to compute the reconstruction loss:
\begin{equation}
    \mathcal{L}_{rec} = ( \mathcal{X} - y^{rec})^2.
    \label{eq:eq3}
\end{equation}

\begin{figure}[t!]
\centering 
\includegraphics[width=.79\textwidth]{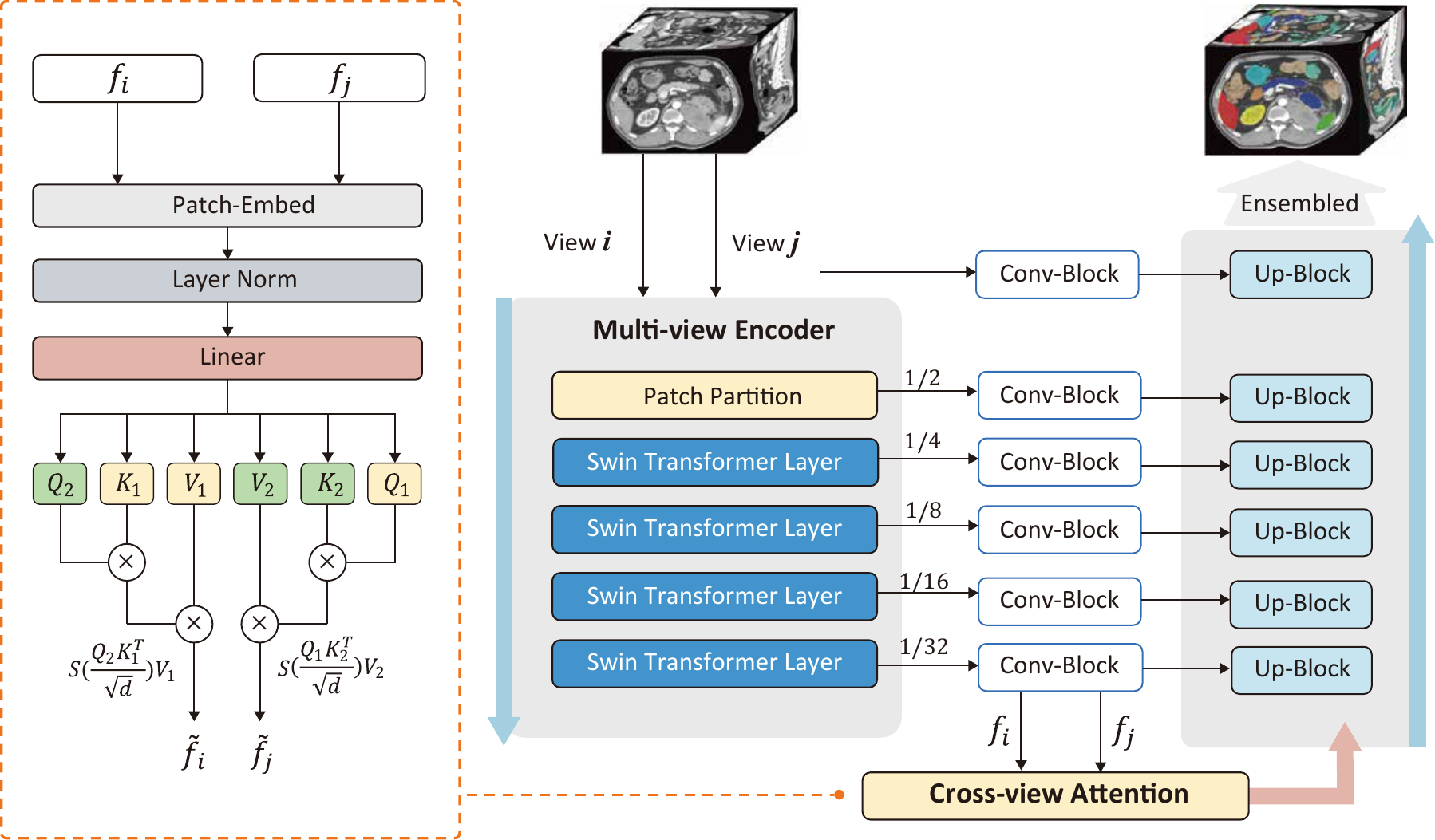}
\caption{Overview of our proposed SwinMM. The Conv-Blocks convolve the latent representations obtained from different levels of the masked multi-view encoder, adapting their feature size to match that of the corresponding decoder layer. The Up-Blocks perform deconvolution to upsample the feature maps.}
\label{framework}
\end{figure}

\item\textbf{The rotation task}  aims to detect the rotation angle of the masked input along the axis of the selected perspective, with possible rotation angles of $0^\circ$, $90^\circ$, $180^\circ$, and $270^\circ$. The model's performance is evaluated using cross-entropy loss, as shown in Eq. \ref{eq:eq4}, where $y^{rot}$ and $y_{r}$ represent the predicted probabilities of the rotation angle and the ground truth, respectively.
\begin{equation}
    \mathcal{L}_{rot} = -\sum^{R}_{r=1} y_r\log(y^{rot}).
\label{eq:eq4}
\end{equation}

\item\textbf{The contrastive learning task} aims to assess the effectiveness of a model in representing input data by comparing high-level features of multiple views. Our working assumption is that although the representations of the same sample may vary at the local level when viewed from different perspectives, they should be consistent at the global level. To compute the contrastive loss, we use cosine similarity $sim(\cdot)$, where $y_i^{con}$ and $y_j^{con}$ represent the contrastive pair, $t$ is a temperature constant, and $1$ is the indicator function.
\begin{equation}
    \mathcal{L}_{con} = 
                -\log\frac{\exp(sim(y_i^{con}, y_j^{con})/t)}{\sum^{2N}_{k}1_{k\neq i}\exp(sim(y_i^{con}, y_k^{con})/t)}.
\label{eq:eq5}
\end{equation}

\item\textbf{The mutual learning task} assesses the consistency of reconstruction results from different views to enable the model to learn aligned information from multi-view inputs. Reconstruction results are transformed into a uniform perspective and used to compute a mutual loss $\mathcal{L}_{mul}$, which, like the reconstruction task, employs the MSE loss. Here, $y^{rec}{i}$ and $y^{rec}{j}$ represent the predicted reconstruction from views $i$ and $j$, respectively.

\begin{equation}
    \mathcal{L}_{mul} = ( y^{rec}_{i} - y^{rec}_{j})^2.
    \label{eq:eq6}
\end{equation}

\end{itemize}

The total pre-training loss is as shown in Eq. \ref{eq:eq7}. The weight coefficients $\alpha_{1}$, $\alpha_{2}$, $\alpha_{3}$ and $\alpha_{4}$ are set equal in our experiment ($\alpha_{1}=\alpha_{2}=\alpha_{3}=\alpha_{4}=1$).
\begin{equation}
    \mathcal{L}_{pre}=\alpha_{1}\mathcal{L}_{rec}+\alpha_{2}\mathcal{L}_{rot}+\alpha_{3}\mathcal{L}_{con}+\alpha_{4}\mathcal{L}_{mul}.
    \label{eq:eq7}
\end{equation}
\subsection{Fine-tuning}
\paragraph{\textbf{Cross-view decoder.}}
\label{cross-view decoder}
The structure of the cross-view decoder, comprising Conv-Blocks for skip connection, Up-Blocks for up-sampling, and a Cross-view Attention block for views interaction, is depicted in Figure \ref{framework}. The Conv-Blocks operate on different layers, reshaping the latent representations from various levels of the masked multi-view encoder by performing the convolution, enabling them to conform to the feature size in corresponding decoder layers ($\frac{H}{2^{i}},\frac{W}{2^{i}},\frac{D}{2^{i}}, i=0,1,2,3,4,5$). At the bottom of the U-shaped structure, the cross-view attention module integrates the information from two views. The representations at this level are assumed to contain similar semantics. The details of the cross-view attention mechanism are presented in Figure \ref{framework} and Eq. \ref{eq-crossattn}. In the equation, $f_i$ and $f_j$ denote the representations of different views, while $Q_i$, $K_i$, and $V_i$ refer to the \emph{query}, \emph{key}, and \emph{value} matrices of $f_i$, respectively.
\begin{align}
\begin{array}{r}
 \text{Cross Attention}(f_i,f_j)
 =[\textnormal{Softmax}\left(\frac{Q_{i}K_{j}^{\top}}{\sqrt{d}}\right)V_{j}, \textnormal{Softmax}\left(\frac{Q_{j}K_{i}^{\top}}{\sqrt{d}}\right)V_{i}].
 \end{array}
 \label{eq-crossattn}
\end{align}

\paragraph{\textbf{Multi-view consistency loss.}}

We assume consistent segmentation results should be achieved across different views of the same volume. To quantify the consistency of the multi-view results, we introduce a consistency loss $\mathcal{L}_{mc}$, calculated using KL divergence in the fine-tuning stage, as in previous work on mutual learning ~\cite{Zhang2018DeepML}. The advantage of KL divergence is that it does not require class labels and has been shown to be more robust during the fine-tuning stage. We evaluate the effectiveness of different mutual loss functions in an ablation study (see supplementary). 
The KL divergence calculation is shown in Eq. \ref{eq3:eq2}:

\begin{equation}
\begin{array}{l}
    \mathcal{L}_{MC} =D_{K L}(V_i \| V_j)=\sum_{m=1}^{N} V_i\left(x_{m}\right) \cdot \log \frac{V_i\left(x_{m}\right)}{V_j\left(x_{m}\right)},              
    \label{eq3:eq2}
\end{array}
\end{equation}
where 
$V_i(x_m)$ and $V_j(x_m)$ denote the different view prediction of m-th voxel. 
$N$ represents the number of voxels of case $x$. $V_i(x)$ and $V_j(x)$ denote different view prediction of case $x$.
We measure segmentation performance using $\mathcal{L}_{DiceCE}$, which combines Dice Loss and Cross Entropy Loss according to ~\cite{tang2022self}. 
\begin{equation}
\begin{array}{r}
    \mathcal{L}_{DiceCE} =
    1-\sum_{m=1}^{N}({\frac{2\left|p_{m}\cap y_{m}\right|}{N\left(\left|p_{m}\right|+\left|y_{m}\right|\right)}}+{\frac{y_{m}\log(p_{m})}{N}}),
    \label{eq:eq1}
\end{array}
\end{equation}
 where $p_{m}$ and $y_{i}$ respectively represent the predicted and ground truth labels for the m-th voxel, while $N$ is the total number of voxels. We used $\mathcal{L}_{fin}$ during the fine-tuning stage, as specified in Eq. \ref{eq:eq2}, and added weight coefficients $\beta_{DiceCE}$ and $\beta_{mc}$ for different loss functions, both set to a default value of 1.
\begin{equation}
\begin{array}{l}
    \mathcal{L}_{fin} =\beta_{DiceCE}\mathcal{L}_{DiceCE}+\beta_{MC}\mathcal{L}_{MC}.
    \label{eq:eq2}
\end{array}
\end{equation}

\subsection{Semi-supervised learning with SwinMM}
As mentioned earlier, the multi-view nature of SwinMM can substantially enhance the reliability and accuracy of its final output while minimizing the need for large, high-quality labeled medical datasets, making it a promising candidate for semi-supervised learning. In this study, we propose a simple variant of SwinMM to handle semi-supervision. As depicted in Figure~\ref{Semi}, we leverage the diverse predictions from different views for unlabeled data and generate aggregated pseudo-labels for the training process. Compared to single-view models, SwinMM's multi-view scheme can alleviate prediction uncertainty by incorporating more comprehensive information from different views, while ensemble operations can mitigate individual bias.

\section{Experiments}

\paragraph{\textbf{Datasets and evaluation.}} Our pre-training dataset includes 5833 volumes from 8 public datasets: AbdomenCT-1K~\cite{ma2021abdomenct}, BTCV~\cite{iglesias2015multi}, MSD~\cite{antonelli2022medical}, TCIA-Covid19~\cite{harmon2020artificial}, WORD~\cite{luo2022word}, TCIA-Colon~\cite{Johnson2009AccuracyOC}, LiDC~\cite{Armato2011TheLI}, and HNSCC~\cite{Grossberg2018ImagingAC}. 
We choose two popular datasets, WORD (The Whole abdominal ORgan Dataset) and ACDC~\cite{Bernard2018DeepLT} (Automated Cardiac Diagnosis Challenge), to test the downstream segmentation performance. The accuracy of our segmentation results is evaluated using two commonly used metrics: the Dice coefficient and Hausdorff Distance (HD).

\paragraph{\textbf{Implementation details.}}
Our SwinMM is trained on 8 A100 Nvidia GPUs with 80G gpu memory. In the pre-training process, we use a masking ratio of 50\%, a batch size of 2 on each GPU, and an initial learning rate of 5e-4 and weight decay of 1e-1. In the finetuning process, we apply a learning rate of 3e-4 and a layer-wise learning rate decay of 0.75. We set 100K steps for pre-training and 2500 epochs for fine-tuning. We use the AdamW optimizer and the cosine learning rate scheduler in all experiments with a warm-up of 50 iterations to train our model. We follow the official data-splitting methods on both WORD and ACDC, and report the results on the test dataset. For inference on these datasets, we applied a double slicing window inference, where the window size is $64\times64\times64$ and the overlapping between windows is 70\%.

\subsection{Results}
\paragraph{\textbf{Comparing with SOTA baselines.}}
We compare the segmentation performance of SwinMM with several popular and prominent networks, comprising fully supervised networks, i.e., U-Net~\cite{ronneberger2015u}, Swin UNet~\cite{liu2021swin}, VT-UNet~\cite{peiris2022robust}, UNETR~\cite{Hatamizadeh2022UNETRTF}, DeepLab V3+~\cite{Chen2018EncoderDecoderWA}, ESPNet~\cite{Mehta2018ESPNetES}, DMFNet~\cite{Chen20193DDM}, and LCOVNet~\cite{Zhao2021LCOVNETAL}, as well as self-supervised method Swin UNETR~\cite{tang2022self}. As shown in Table~\ref{ACDC} and Table~\ref{Baseline}, our proposed SwinMM exhibits remarkable efficiency in medical segmentation by surpassing all the other pipelines and achieves higher average Dice (86.18\% on WORD and 90.80\% on ACDC) and lower HD (9.35 on WORD and 6.37 on ACDC). 

\paragraph{\textbf{Single view vs. multiple views.}} To evaluate the effectiveness of our proposed multi-view self-supervised pretraining pipeline, we compared it with the state-of-the-art self-supervised learning method SwinUNETR~\cite{tang2022self} on  WORD~\cite{luo2022word} dataset. Specifically,  two SwinUNETR-based methods are compared: using fixed single views (Axial, Sagittal, and Coronal) and using ensembled predictions from multiple views (denoted as SwinUNETR-Fuse). Our results, presented in Table \ref{WORD}, show that our SwinMM surpasses all other methods including SwinUNETR-Fuse, highlighting the advantages of our unique multi-view designs. Moreover, by incorporating multi-view ensemble operations, SwinMM can effectively diminish the outliers in hard labels and produce more precise outputs, especially when dealing with harder cases such as smaller organs. The supplementary material provides qualitative comparisons of 2D/3D segmentation outcomes.

\begin{table*}[!ht]
\centering
\captionsetup{width=\textwidth}
\caption{Quantitative results of ACDC dataset. Note: RV - right ventricle, Myo - myocardium, LV - left ventricle.}
\resizebox{0.68\linewidth}{!}{
\begin{tabular}{m{4cm}|cccc|cccc}
\toprule
\textbf{Methods} & \multicolumn{4}{c|}{DICE (\%) $\uparrow$} & \multicolumn{4}{c}{HD $\downarrow$} \\
 & RV & Myo & LV & Average & RV & Myo & LV & Average \\
\midrule
U-Net~\cite{ronneberger2015u} &54.17 &43.92 &60.23 &52.77 &24.15 &35.32 &60.16 &39.88\\
Swin UNet~\cite{liu2021swin} &78.50 &77.92 &86.31 &80.91 & 11.42 & 5.61 &7.42 &8.12\\
VT-UNet~\cite{peiris2022robust} &80.44 &80.71 &89.53 &83.56 &11.09 &5.24 &6.32 &7.55\\
UNETR~\cite{Hatamizadeh2022UNETRTF} &84.52 &84.36 &92.57 &87.15 &12.14 &5.19 &4.55 &7.29\\
Swin UNETR~\cite{tang2022self} & 87.49 & 88.25 & 92.72 & 89.49 & 12.45 & 5.78 & 4.03 & 7.42 \\
\midrule
\rowcolor{mygray}
\textbf{SwinMM} & 90.21 & 88.92 & 93.28 & 90.80 & 8.85 & 3.10 & 7.16 & 6.37 \\
\bottomrule
\end{tabular}
}
\label{ACDC}
\end{table*}
\begin{table*}[!ht]
\caption{Quantitative results of WORD dataset. Note: Liv - liver, Spl - spleen, Kid L - left kidney, Kid R - right kidney, Sto - stomach, Gal - gallbladder, Eso - esophagus, Pan - pancreas, Duo - duodenum, Col - colon, Int - intestine, Adr - adrenal, Rec - rectum, Bla - bladder, Fem L - left femur, Fem R - right femur.}
\resizebox{\linewidth}{!}{
\centering
\begin{tabular}{l|cccccccccccccccccc}
\toprule
\textbf{Methods} & Liv & Spl & Kid L & Kid R & Sto & Gal & Eso & Pan & Duo & Col & Int & Adr & Rec & Bla & Fem L & Fem R & \textbf{DICE(\%)}$\uparrow$ & \textbf{HD}$\downarrow$ \\ 
\midrule
UNETR~\cite{Hatamizadeh2022UNETRTF} & 94.67 & 92.85 & 91.49 & 91.72 & 85.56 & 65.08 & 67.71 & 74.79 & 57.56 & 74.62 & 80.4 & 60.76 & 74.06 & 85.42 & 89.47 & 90.17 & 79.77 & 17.34 \\
CoTr~\cite{Xie2021CoTrEB} & 95.58 & 94.9 & 93.26 & 93.63 & 89.99 & 76.4 & 74.37 & 81.02 & 63.58 & 84.14 & 86.39 & 69.06 & 80.0 & 89.27 & 91.03 & 91.87 & 84.66 & 12.83 \\
DeepLab V3+~\cite{Chen2018EncoderDecoderWA} & 96.21 & 94.68 & 92.01 & 91.84 & 91.16 & 80.05 & 74.88 & 82.39 & 62.81 & 82.72 & 85.96 & 66.82 & 81.85 & 90.86 & 92.01 & 92.29 & 84.91 & 9.67 \\
Swin UNETR~\cite{tang2022self} & 96.08 & 95.32 & 94.20 & 94.00 & 90.32 & 74.86 & 76.57 & 82.60 & 65.37 & 84.56 & 87.37 & 66.84 & 79.66 & 92.05 & 86.40 & 83.31 & 84.34& 14.24\\
\midrule
ESPNet~\cite{Mehta2018ESPNetES} & 95.64 & 93.9 & 92.24 & 94.39 & 87.37 & 67.19 & 67.91 & 75.78 & 62.03 & 78.77 & 72.8 & 60.55 & 74.32 & 78.58 & 88.24 & 89.04 & 79.92 & \multicolumn{1}{c}{15.02} \\
DMFNet~\cite{Chen20193DDM} & 95.96 & 94.64 & 94.7 & 94.96 & 89.88 & 79.84 & 74.1 & 81.66 & 66.66 & 83.51 & 86.95 & 66.73 & 79.26 & 88.18 & 91.99 & 92.55 & 85.1 & \multicolumn{1}{c}{7.52} \\
LCOVNet~\cite{Zhao2021LCOVNETAL} & 95.89 & 95.4 & 95.17 & 95.78 & 90.86 & 78.87 & 74.55 & 82.59 & 68.23 & 84.22 & 87.19 & 69.82 & 79.99 & 88.18 & 92.48 & 93.23 & 85.82 & \multicolumn{1}{c}{9.11} \\
\midrule
\rowcolor{mygray}
\textbf{SwinMM} & 96.30 & 95.46 & 93.83 & 94.47 & 91.43 & 80.08 & 76.59 & 83.60 & 67.38 & 86.42 & 88.58 & 69.12 & 80.48 & 90.56 & 92.16 & 92.40 & 86.18 & 9.35 \\
\bottomrule
\end{tabular}}
\label{Baseline}
\end{table*}

\begin{table*}[!ht]
\caption{Quantitative results of WORD dataset. Abbreviations follows Table~\ref{Baseline}.}
\resizebox{\linewidth}{!}{
\centering
\begin{tabular}{l|cccccccccccccccccc}
\toprule
\textbf{Methods} & Liv & Spl & Kid L & Kid R & Sto & Gal & Eso & Pan & Duo & Col & Int & Adr & Rec & Bla & Fem L & Fem R & \textbf{DICE(\%)}$\uparrow$ & \textbf{HD}$\downarrow$ \\ 
\midrule
Swin UNETR Axi & 96.08 & 95.32 & 94.20 & 94.00 & 90.32 & 74.86 & 76.57 & 82.60 & 65.37 & 84.56 & 87.37 & 66.84 & 79.66 & 92.05 & 86.40 & 83.31 & 84.34& 14.24 \\
Swin UNETR Sag & 96.09 & 95.32 & 93.53 & 94.72 & 90.68 & 73.31 & 74.10 & 83.07 & 66.98 & 84.21 & 86.37 & 68.07 & 78.89 & 91.18 & 91.67 & 91.28 & 84.97 & 40.88\\
Swin UNETR Cor & 96.12 & 95.49 & 93.91 & 94.80 & 90.25 & 71.78 & 75.27 & 82.83 & 66.26 & 84.07 & 86.98 & 66.23 & 79.38 & 90.93 & 88.09 & 86.74 & 84.32 & 14.02\\
Swin UNETR Fuse & 96.25 & 95.71 & 94.20 & 94.85 & 91.05 & 74.80 & 77.04 & 83.73 & 67.36 & 85.15 & 87.69 & 67.84 & 80.29 & 92.31 & 90.44 & 89.36 & 85.50 & 13.87\\
\midrule
\rowcolor{mygray}
\textbf{SwinMM} & 96.30 & 95.46 & 93.83 & 94.47 & 91.43 & 80.08 & 76.59 & 83.60 & 67.38 & 86.42 & 88.58 & 69.12 & 80.48 & 90.56 & 92.16 & 92.40 & 86.18 & 9.35 \\
\bottomrule
\end{tabular}}
\label{WORD}
\end{table*}

\begin{table}[!ht]
    \centering
    \begin{minipage}[t]
    {0.5\linewidth}
        \centering
        \caption{
        The ablation study of proxy tasks during pre-training.
        }
        \resizebox{\linewidth}{!}{
            \begin{tabular}
            {l|cccccc}
            \toprule
            \textbf{Methods} & Rec & Mut & Rot & Con & DICE(\%) $\uparrow$ & HD $\downarrow$ \\ 
            \midrule
            SwinMM (w/o pretraining) & -- & -- & -- & -- & 84.78 & 11.77 \\
            SwinMM  & \checkmark &--  &--  &--  & 84.93& 11.61 \\
            SwinMM & -- & -- & \checkmark & \checkmark & 84.91 & 11.69 \\
            SwinMM  & \checkmark & \checkmark & -- & --  &  85.65& 10.25 \\
            SwinMM & \checkmark & -- & \checkmark & \checkmark & 85.19 & 10.98 \\
            \rowcolor{mygray}
            SwinMM & \checkmark & \checkmark & \checkmark & \checkmark & 85.74 & 9.56 \\ \bottomrule
            \end{tabular}}
        \label{tab:ablation_pre_train_function}
    \end{minipage}
    \hfill
    \begin{minipage}[t]{0.48\linewidth}
        \captionsetup{width=0.7\textwidth}
        \caption{
        The ablation study of label ratios. 
        }
	\centering
            \resizebox{0.6\linewidth}{!}{
            \begin{tabular}{l|cc}
            \toprule
            \textbf{label ratio} &Swin UNETR & SwinMM  \\ 
            \midrule
             10\%& 56.14& 67.83 
             \\
            30\%& 70.65& 78.91 
            \\
            50\%& 77.28&82.03 
            \\
            70\%& 81.07&83.25  
            \\
            90\%& 82.46 &84.32 
            \\ 
            \rowcolor{mygray}
            100\%& 83.13 &84.78 
            \\ 
            \bottomrule
            \end{tabular}}
        \label{ratio_label}
    \end{minipage}
\end{table}

\subsection{Ablation Study}
To fairly evaluate the benefits of our proposed multi-view design, we separately investigate its impact in the pre-training stage, the fine-tuning stage, as well as both stages. Additionally, we analyze the role of each pre-training loss functions. 
\paragraph{\textbf{Pre-training loss functions.} }
The multi-view pre-training is implemented by proxy tasks. The role of each task can be revealed by taking off other loss functions. For cheaper computations, we only pre-train our model on 2639 volumes from 5 datasets (AbdomenCT-1K, BTCV, MSD, TCIA-Covid19, and WORD) in these experiments, and we applied a 50\% overlapping window ratio, during testing time. As shown in Table~\ref{tab:ablation_pre_train_function}, our proposed mutual loss brings a noticeable improvement in Dice (around 1\%) over the original SwinUNETR setting. When combining all the proxy tasks, our SwinMM achieves the best performance.

\paragraph{\textbf{Data efficiency.}} The data efficiency is evaluated under various semi-supervised settings. Initially, a base model is trained from scratch with a proportion of supervised data from the WORD dataset for 100 epochs. Then, the base model finishes the remaining training procedure with unsupervised data. The proportion of supervised data (denoted as label ratio) varies from 10\% to 100\%. Table~\ref{ratio_label} shows SwinMM consistently achieves higher Dice (\%) than SwinUNETR, and its superiority is more remarkable when training with fewer supervised data.

\section{Conclusion}
This paper introduces SwinMM, a self-supervised multi-view pipeline for medical image analysis. SwinMM integrates a masked multi-view encoder in the pre-training phase and a cross-view decoder in the fine-tuning phase, enabling seamless integration of multi-view information, thus boosting model accuracy and data efficiency. Notably, it introduces a new proxy task employing a mutual learning paradigm, extracting hidden multi-view information from 3D medical data. The approach achieves competitive segmentation performance and higher data-efficiency than existing methods and underscores the potential and efficacy of multi-view learning within the domain of self-supervised learning.

\paragraph{\textbf{Acknowledgement.}}
This work is partially supported by the Google Cloud Research Credits program.

\bibliographystyle{splncs04}
\bibliography{egbib}

\newpage

\begin{table}[ht]
	\centering
         \caption{
            The ablation study of the pre-training masking ratio. We fine-tune our pre-trained model on the WORD dataset. The result shows that an average masking ratio (50\%) helps the network achieve higher Dice and lower HD. 
            }
        \resizebox{0.7\columnwidth}{!}{%
            \begin{tabular}{l|ccc}
            \toprule
            \textbf{} & Masking ratio & DICE(\%) $\uparrow$& HD(\%)$\downarrow$ \\
            \midrule
            \midrule
            SwinMM w/o. Pretrain & -- & 84.78\% & 11.77 \\
            SwinMM & 0.25 & 85.54\% & 12.34  \\\rowcolor{mygray}
            SwinMM & 0.50 &85.74\% & 9.56 \\
            SwinMM & 0.75 &85.43\% &10.50\\
            \bottomrule
            \end{tabular}}
        \label{mask_ratio}    
\end{table}

\begin{table}[ht]
    \centering
    \caption{
            The ablation study of the multi-view components in fine-tuning. Multi-view Consistency Loss Different loss functions are investigated for the multi-view consistency computation. The results indicate that both KL Loss and Cosine Loss improve the Average Dice, and KL Loss performs better than Cosine Loss. "Bottleneck" represents implementing a Cross Attention layer between the deepest Conv-Block and Up-Block. "Intermediate Layer" means the Cross Attention layer is at a shallower position. The results demonstrate the advantage of adding the cross-view attention block in bottleneck, and deeper layer representations contain higher-level information that is more valuable.}
        \begin{tabular}{l|cccc}
            \toprule
            \textbf{Methods} & Consistency Loss & Cross Attn layer & DICE(\%) $\uparrow$ & HD $\downarrow$ \\ \midrule
            \midrule
            SwinMM & -- & -- & 83.72\% & 16.24 \\
            SwinMM & KL & -- & 84.44\% & 13.52 \\
            SwinMM & -- & Bottleneck & 84.49\% & 13.68 \\
            SwinMM & Cosine & Bottleneck & 84.54\% & 13.22 \\
            SwinMM & KL & Intermediate Layer & 84.55\% & 12.16 \\
            \rowcolor{mygray}
            SwinMM & KL & Bottleneck & \textbf{84.78\%} & \textbf{11.77} \\
            \bottomrule
            \end{tabular}
\end{table}

\clearpage

\begin{figure}
  \centering
  \includegraphics[width=0.53\linewidth]{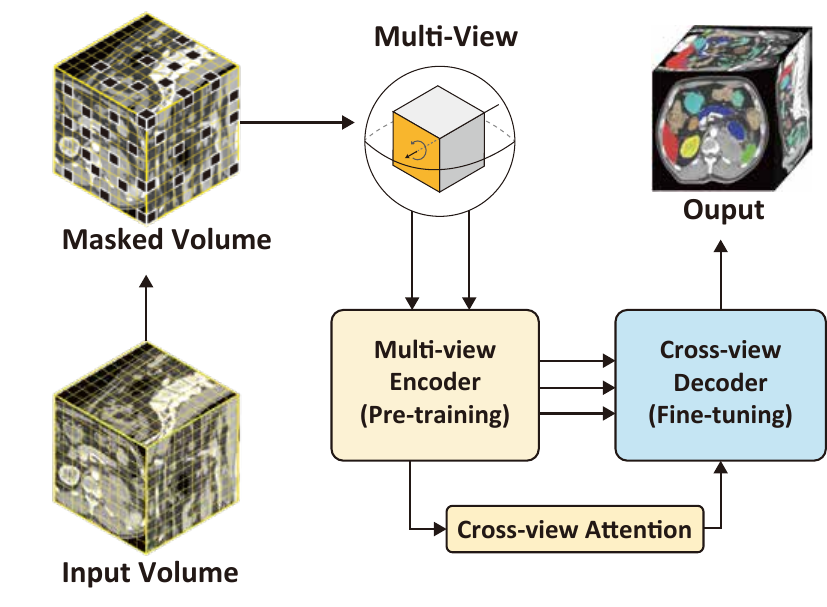}
  \caption{The structure of SwinMM. 
  }
  \label{fig:abstract_fig}
  \setlength{\belowcaptionskip}{-1cm}
\end{figure}

\begin{figure}[!htb]
\centering 
\includegraphics[width=\textwidth]{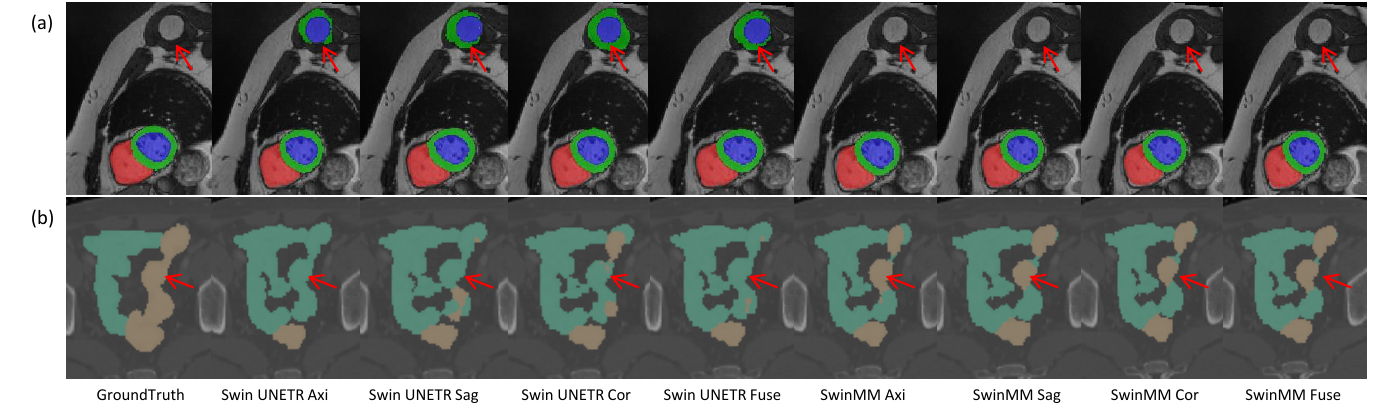}
\caption{The 2D qualitative comparison between our method (SwinMM) and baseline method (Swin UNETR ~\cite{tang2022self}) on the  ACDC (a) and the WORD dataset (b).}
\label{Visualized_Result}
\end{figure}

\textbf{\begin{figure}[!htb]
\centering 
\includegraphics[width=\textwidth]{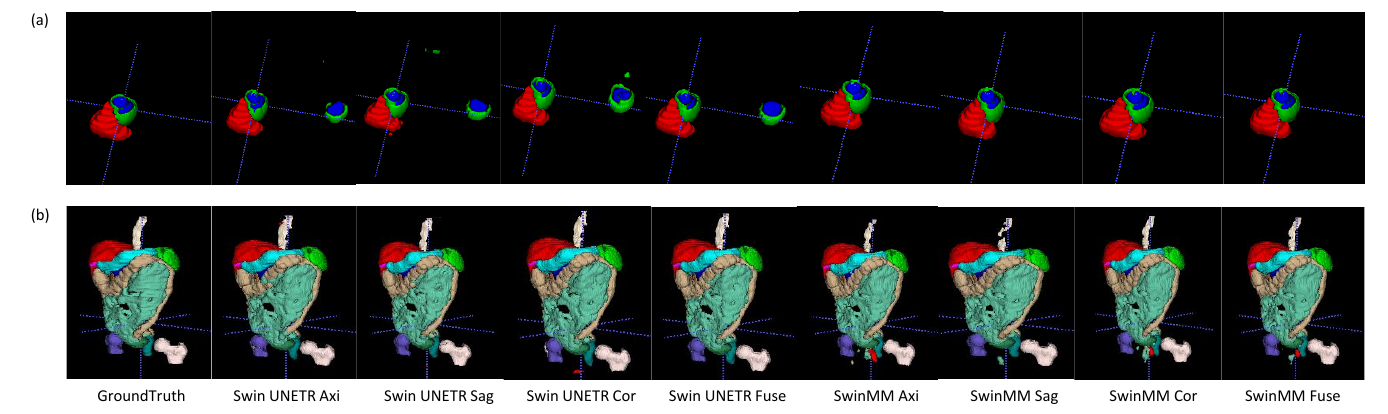}
\caption{The 3D qualitative comparison between our method (SwinMM) and baseline method (Swin UNETR ~\cite{tang2022self}) on the  ACDC (a) and the WORD dataset (b).}
\label{Visualized_Result_2}
\end{figure}}
\end{document}